%% file: main.tex
\documentclass[11pt]{article}

\usepackage{times}
\usepackage{latexsym}
\usepackage{booktabs}
\usepackage{amsfonts}
\usepackage{nicefrac}
\usepackage{microtype}
\usepackage{todonotes}
\usepackage{listings}
\usepackage{algorithm}
\usepackage{algpseudocode}
\usepackage{soul}
\usepackage{tabularx}
\usepackage{pifont}

\usepackage{xcolor}
\usepackage{amsmath}
\usepackage{hyperref}
\usepackage{cleveref}
\usepackage{comment}
\usepackage{tcolorbox}
\usepackage{nicematrix}

\usepackage{colortbl}
\usepackage{makecell}

\usepackage[normalem]{ulem}

\input{macro}

\usepackage[T1]{fontenc}

\usepackage[utf8]{inputenc}

\usepackage{microtype}
\usepackage{algpseudocode,algorithm}

\usepackage{multirow}
\usepackage{graphicx}
\usepackage{pdfpages}
\usepackage{url}
\usepackage{epsfig}
\usepackage{adjustbox}
\usepackage{amsfonts}

\usepackage{amssymb}
\usepackage{booktabs}
\usepackage{comment}
\usepackage{arydshln}
\usepackage{caption}
\usepackage{microtype}
\usepackage{float}

\usepackage{xspace}
\usepackage[final]{acl}
\usepackage{kotex}
\usepackage{setspace}
\title{YA-TA: Towards Personalized Question-Answering Teaching Assistants using Instructor-Student Dual Retrieval-augmented Knowledge Fusion}

\author{
  \textbf{Dongil Yang}\thanks{~~Equal contribution} \qquad   \textbf{Suyeon Lee}\footnotemark[1] \qquad  \textbf{Minjin Kim} \\ \qquad \textbf{Jungsoo Won}\qquad \textbf{Namyoung Kim}\qquad 
\textbf{Dongha Lee}\thanks{\ \ Co-corresponding authors}\qquad 
\textbf{Jinyoung Yeo}\footnotemark[2]\\ 
Yonsei University  \\
\texttt{\{wingu, isuy.groot, donalee, jinyeo\}@yonsei.ac.kr}\\
}

\begin{document}
\maketitle
\input{0_abs}
\input{1_intro_new}
\input{3_method}
\input{5_exp_new}
\input{5_results}
\input{2_rel}
\input{6_conc}

\bibliography{anthology,custom}
\bibliographystyle{acl_natbib}

\newpage
\appendix
\input{8_appendix}

\end{document}

%% file: macro.tex
\crefformat{section}{\S#2#1#3}
\crefformat{subsection}{\S#2#1#3}
\crefformat{subsubsection}{\S#2#1#3}
\crefrangeformat{section}{\S\S#3#1#4 to~#5#2#6}
\crefmultiformat{section}{\S\S#2#1#3}{ and~#2#1#3}{, #2#1#3}{ and~#2#1#3}

\definecolor{mypink1}{rgb}{0.858, 0.188, 0.478}

\definecolor{olive}{rgb}{0.39, 0.875, 0.12}

\newcommand{\eg}{{\it e.g.}}%
\newcommand{\ie}{{\it i.e.}}%
\newcommand{\etc}{{\it etc.}}%

\definecolor{yellow-green}{rgb}{0.3, 0.5, 0.0}

\newcommand{\ours}{\textsc{DRaKe}\xspace}

\definecolor{lightblue}{RGB}{224,236,247}
\definecolor{deepblue}{RGB}{9,46,107}
\definecolor{defaultcolor}{gray}{0.9}

\newcommand{\info}{\raisebox{-2pt}{\includegraphics[width=1em]{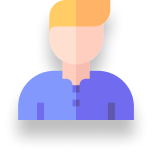}}}
\newcommand{\query}{\includegraphics[width=0.8em]{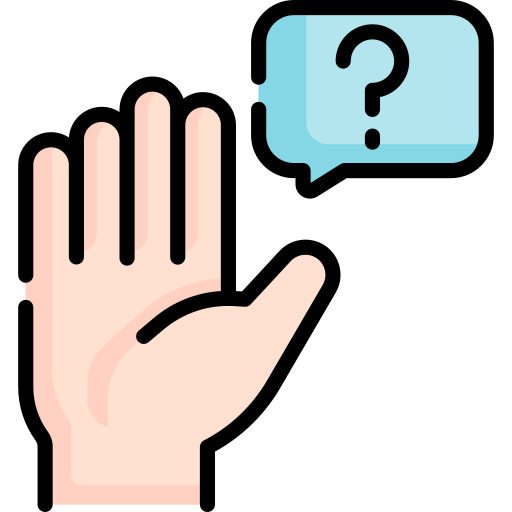}}
\newcommand{\evidence}{\includegraphics[width=0.7em]{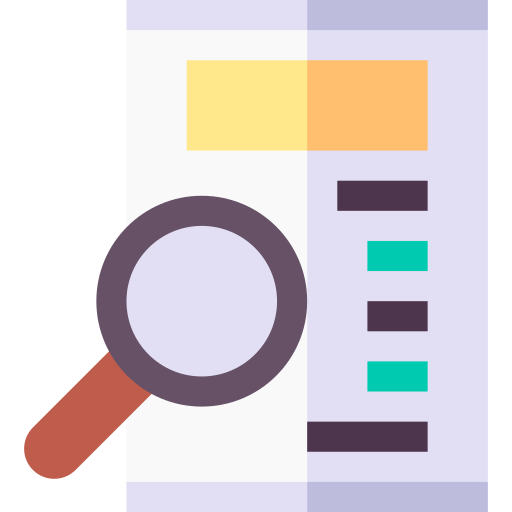}}
\newcommand{\YATAresponse}{\includegraphics[width=1em]{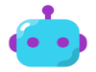}}
\newcommand{\GPTresponse}{\raisebox{-2pt}{\includegraphics[width=1em]{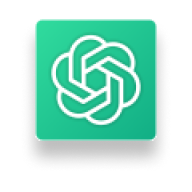}}}

%% file: 0_abs.tex
\begin{abstract}
Engagement between instructors and students plays a crucial role in enhancing students' academic performance. However, instructors often struggle to provide timely and personalized support in large classes. To address this challenge, we propose a novel Virtual Teaching Assistant (VTA) named YA-TA, designed to offer responses to students that are grounded in lectures and are easy to understand. To facilitate YA-TA, we introduce the Dual Retrieval-augmented Knowledge Fusion (\ours{}) framework, which incorporates dual retrieval of instructor and student knowledge and knowledge fusion for tailored response generation. Experiments conducted in real-world classroom settings demonstrate that the \ours{} framework excels in aligning responses with knowledge retrieved from both instructor and student sides. Furthermore, we offer additional extensions of YA-TA, such as a Q\&A board and self-practice tools to enhance the overall learning experience. Our video is publicly available.\footnote{
Video: \url{https://youtu.be/y2EucPEUgZc}}
\end{abstract}

%% file: 1_intro_new.tex
\section{Introduction}

\input{figures/figure_1}

Active interaction between instructors and students, including tailored feedback to student questions, significantly enhances academic performance~\citep{agwu2023students}. However, when an instructor is responsible for a larger number of students, providing personalized responses to every query becomes challenging. Although Teaching Assistants (TAs) are often employed to address this issue, they frequently struggle to offer timely and personalized responses, while consuming significant manpower and resources~\citep{hicke2023aita}.

This situation underscores the pressing demand for Virtual Teaching Assistants (VTAs) capable of providing personalized tutoring unrestricted by time or location. Large Language Models (LLMs) have demonstrated remarkable conversational capabilities, making LLM-powered TAs well-suited to serve as effective VTAs~\citep{chen2023empowering}. VTAs have two key objectives to enhance educational effectiveness: 1) enhancing instructor convenience by responding to students’ questions in a manner that aligns with the instructor’s teaching style (\ie, instructor-side personalization) and 2) assisting student learning by offering tailored support (\ie, student-side personalization)~\citep{nationalcollege}.

\paragraph{Instructor-side personalization.} To enhance instructor convenience, instructors must find the TA reliable and satisfactory, allowing them to delegate Q\&A tasks confidently. To meet this end, the TA's responses should align with the instructor's lecture, ensuring no conflict between the instructor's explanations and the TA's response.

\paragraph{Student-side personalization.} To support student learning effectively, the TA must provide answers that match students' comprehension levels, helping them deepen their understanding of the course. Given the various academic backgrounds among students, the TA should assess each student's knowledge based on their information and tailor responses accordingly.

There have been several efforts to construct VTAs.~\citet{dong2023build, matsudalangchain} aim to build instructor-personalized VTAs by utilizing Retrieval Augmented Generation (RAG) to generate answers based on external course materials. However, these works do not consider that students' understanding level of the course varies due to different academic backgrounds. On the other hand, ~\citet{park2024empowering} construct student-personalized VTAs by providing responses that consider students' learning styles. However, they do not base their responses on the instructor's teaching style.

To consider both sides of personalization, we introduce YA-TA (Yonsei Academic Teaching Assistant), which, to the best of our knowledge, is the first multi-turn question-answering (QA) agent that incorporates personalization for both instructors and students. However, achieving personalization for both sides is challenging, as it requires integrating information from multiple sources.

To tackle this challenge, we propose \ours{} (\textbf{\underline{D}}ual \textbf{\underline{R}}etrieval-\textbf{\underline{a}}ugmented \textbf{\underline{K}}nowledg\textbf{\underline{e}} Fusion) framework, which consists of two steps before response generation: 1) Dual Retrieval and 2) Knowledge Fusion by integrating retrieved knowledge. On the instructor side, we retrieve the instructor's statements related to the student's query. On the student side, we retrieve the academic information about the student, such as the courses the student has previously taken and their grades. Subsequently, we leverage LLMs' Chain-of-Thought (CoT) abilities~\citep{wei2022chain} to reason over the retrieved knowledge from both sides and generate responses by blending this knowledge.

This approach ensures that the responses align with the instructor's philosophy and are adapted to the student's background. To demonstrate the effectiveness of our method, we conduct experiments on real-world classes. Evaluation results and case study demonstrate that our \ours{} framework significantly enhances personalization for both instructors and students. Additionally, we offer extensions like a Q\&A Board and Self-Practice, which further enrich the student's learning experience. 

%% file: figures/figure_1.tex
\begin{figure}[t!]
    \centering
    \includegraphics[width=0.98\linewidth]{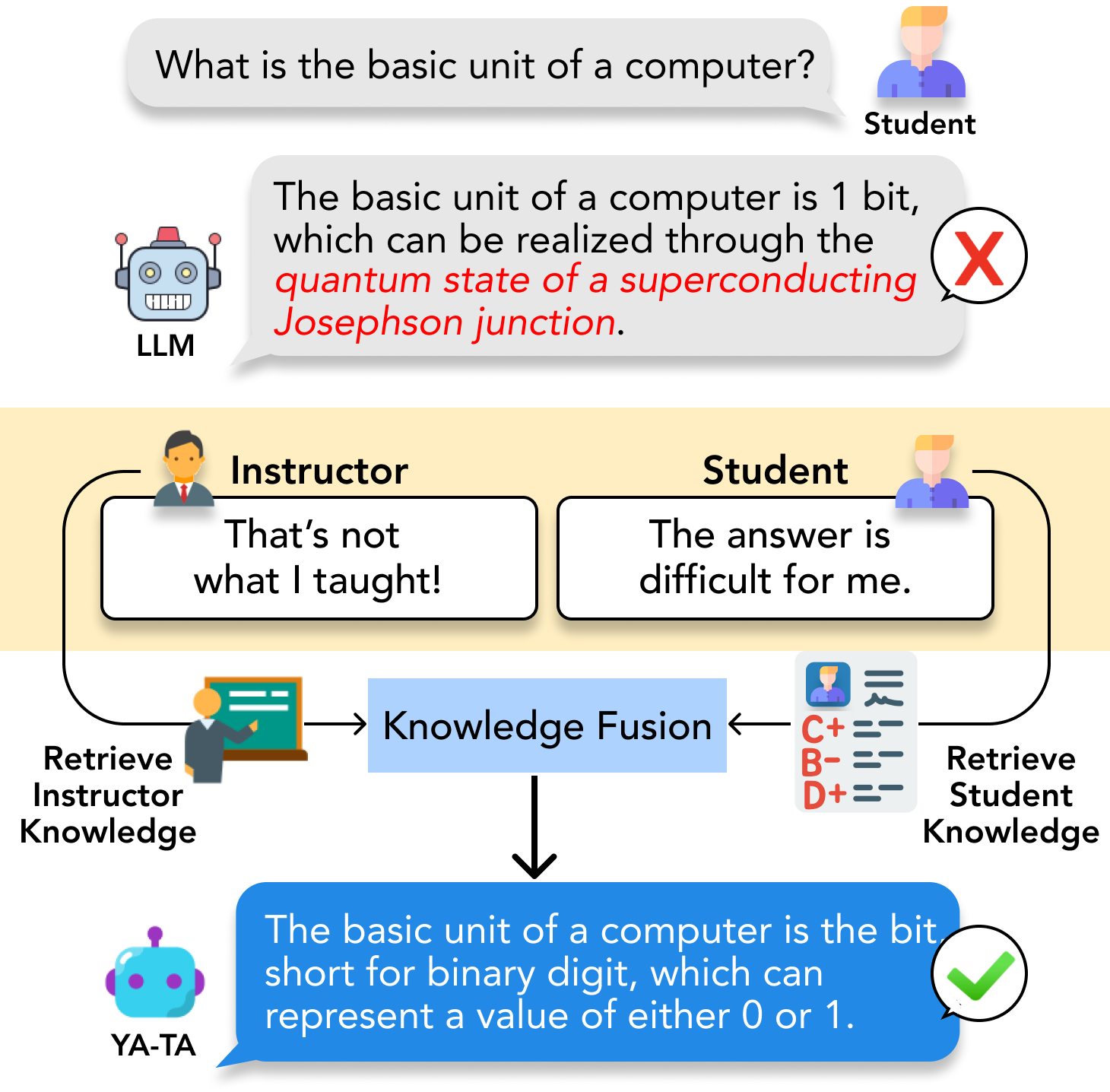}
    \caption{The motivating example of YA-TA. A typical LLM faces challenges in providing responses that consider both instructor and student sides. YA-TA addresses these issues by employing a \ours{} framework.}
    \label{fig:motivation}
\end{figure}



%% file: 3_method.tex
\section{YA-TA}
\input{figures/figure_2}
YA-TA is a multi-turn QA system that aims to generate a reliable and comprehensible response to a student's query. Formally, given the instructor-side knowledge $K_I$, the student-side knowledge $K_S$, and the dialogue context $D_t = \{q_1, r_1, ..., r_{t-1}, q_t\}$ which ends with the student's query, YA-TA's goal is to generate a response $r_t$:

\begin{equation}
    r_t = f(D_t, K_I, K_S)
\end{equation}

To achieve this, we propose Dual Retrieval-augmented Knowledge Fusion (\ours{}), which 1) concurrently retrieves $K_I$ and $K_S$ and 2) integrates them in the response via knowledge fusion module $f(\cdot)$. We explain the data setup process~(\cref{sec:data_setup}), \ours{} framework~(\cref{sec:framework}), and the user interface~(\cref{sec:user_interface}) in this section. The overview of YA-TA is illustrated in Figure~\ref{fig:overview}.

\subsection{Data Setup}
\label{sec:data_setup}
\paragraph{Instructor-side data.}
As YA-TA is based on LLMs, we must handle all data in textual form. When the instructor uploads lecture videos of a course to our system, we extract the audio and run an off-the-shelf automatic speech recognition model to transcribe it into textual segments.\footnote{\url{https://huggingface.co/openai/whisper-large-v3}} Each segment contains transcribed text along with its corresponding timestamp indicating the start and end times of the audio or the video. We store the video and text segments for each lecture in the \textit{instructor course database}, organized by course ID.

\paragraph{Student-side data.}
For the student-side data, we utilize an \textit{academic information system} that contains students' transcripts and a \textit{student query database} that stores past queries. A transcript includes the student's name, major, semester, and grades (\eg\xspace $A^{+}$, $B^{-}$, \etc) of all courses enrolled in the past and represents his or her overall academic performance. While we may use a system from an actual institution, we manually construct a number of transcripts for demonstration purposes. \textit{Student query database} contains a collection of query records of a student. Each record contains queries submitted by the student about a specific course across multiple sessions. Such record indicates the student's comprehension level within the scope of the course.

\subsection{\ours{} Framework}
\label{sec:framework}
\paragraph{Dual retrieval.}
We retrieve knowledge from both instructor and student sides to equip YA-TA with resources to generate reliable and comprehensive responses.
In the \textit{instructor knowledge} retrieval step, we first fetch segments corresponding to a particular course from the \textit{instructor course database}. Among them, we select top $k$ segments by ensembling a sparse and a dense retriever to account for both lexical and semantic similarities when forming $K_I$.\footnote{Implementation details are provided in Appendix~\ref{sec:appendix_yata_details}.} In the \textit{student knowledge} retrieval step, we fetch a student's transcript from the \textit{academic information system} by the student ID and query record from the \textit{student query database} by the course ID. We combine them to form $K_S$ to encompass both the overall and course-specific performances of the student. For example, as shown in Figure~\ref{fig:overview}, when a student named Kelly asks a question about course CS50, YA-TA retrieves a number of relevant segments from CS50 as the \textit{instructor knowledge}, along with Kelly's transcript and query record as the \textit{student knowledge}.

\input{tables/main_table}
\paragraph{Knowledge fusion.}
The main goal of the knowledge fusion module $f(\cdot)$ is to generate a reliable and comprehensible response to the query by integrating \textit{instructor knowledge} and \textit{student knowledge}. However, simply injecting them into the response generator may not produce the best response as each knowledge is composed of raw data. Therefore, we abstract each knowledge to a higher level utilizing the reasoning ability of an LLM, before passing it to the response generator. For the \textit{instructor knowledge}, we use an LLM to extract useful segments as \textit{evidence} that provide necessary information to answer the query. We extract rather than interpret to minimize any deviation from the instructor's exact words, which reflect their principles regarding the course topic. As the \textit{student knowledge} contains raw information such as a transcript or a query record (\ie, list of past queries), we use an LLM to transform it into a \textit{plan}. This \textit{plan} serves as a helpful guide for the response generator, enabling it to personalize the response for the student. Finally, we feed \textit{evidence} and \textit{plan} into the response generator, which then effectively integrates them to produce a response that is both grounded in the lecture and tailored to the student. As illustrated in Figure~\ref{fig:overview}, YA-TA's response, processed through the \ours{} framework, demonstrates a seamless fusion of \textit{instructor knowledge} and \textit{student knowledge}.

\subsection{User Interface: Video Referencing}
\label{sec:user_interface}
As a system designed to assist students learn effectively, we not only offer personalized responses but also enhance the overall learning experience through the user interface. Below the response, we embed the lecture video paused at the exact timestamp of the referenced segment. By replaying the video, students can grasp the full context of the instructor's lecture regarding the response. We control this interface using the title of the lecture video and the timestamp of the referenced segment, both of which are seamlessly generated during the decoding process of the response generator.

%% file: figures/figure_2.tex
\begin{figure*}[t!]
    \centering
    \includegraphics[width=0.98\linewidth]{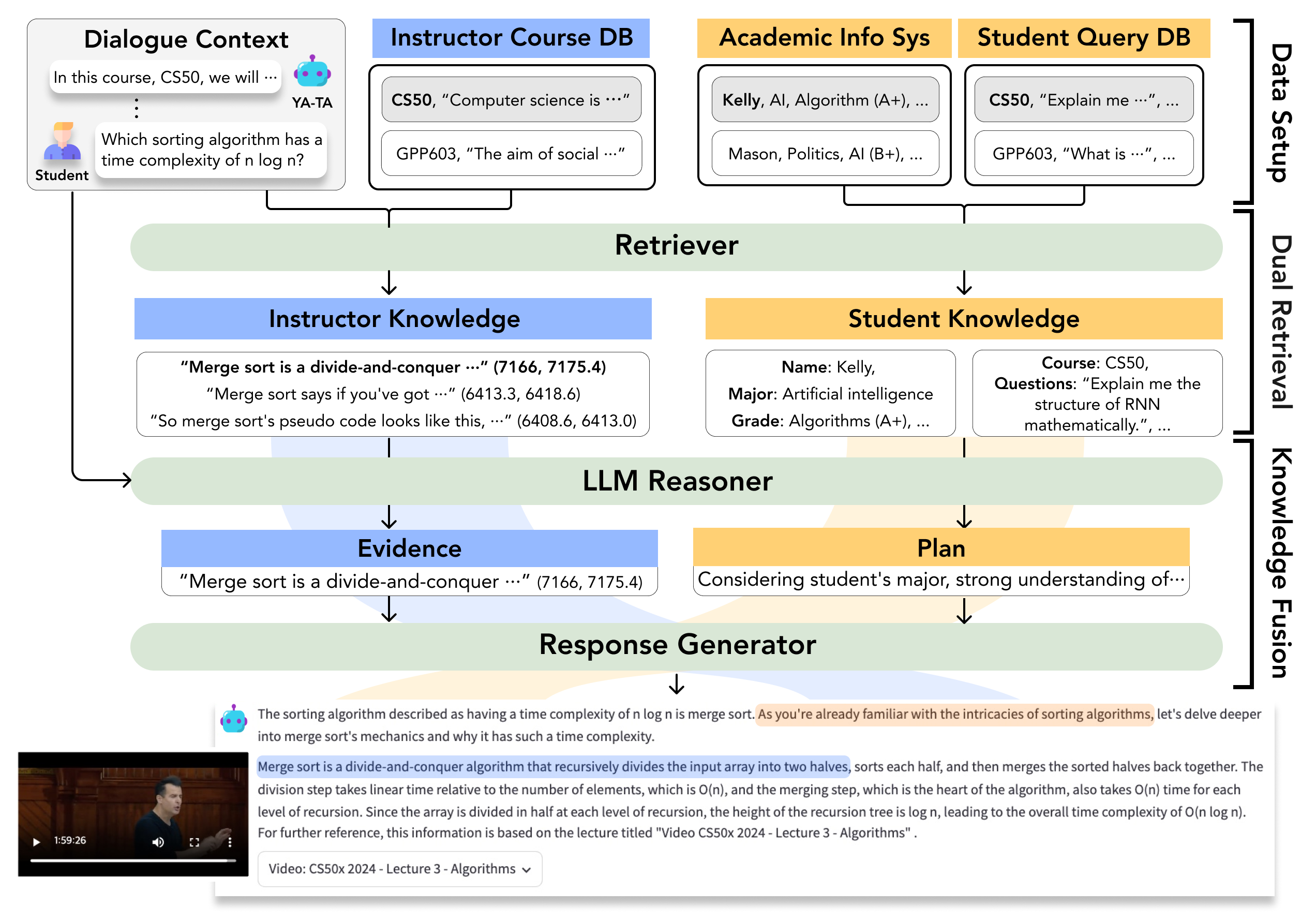}
    \caption{The overview architecture of YA-TA. The image of the response is a screenshot of the YA-TA user interface. In YA-TA's final response, the part highlighted in \textcolor[HTML]{1E90FF}{blue} indicates where instructor-side personalization is evident, while the part highlighted in \textcolor[HTML]{FF8C00}{orange} indicates areas where student-side personalization is evident.}
    \label{fig:overview}
\end{figure*}

%% file: tables/main_table.tex
\begin{table*}[ht!]
\centering
\resizebox{0.98\textwidth}{!}
{
\begin{tabular}{llcccccc}
\toprule
\multirow{2.5}{*}{Model} & \multirow{2.5}{*}{Method} & \multicolumn{2}{c}{Criteria for {Instructor Pers.}} & \multicolumn{2}{c}{Criteria for Student Pers.} & \multicolumn{1}{c}{Criterion for Both Pers.} \\
\cmidrule(lr){3-7} & & Precision & Groundedness & Helpfulness & Comprehensiveness & Overall \\
\midrule
GPT-3.5-Turbo & - & 3.82 & 4.08 & 4.48 & 4.02 & 3.48\\
GPT-4o & - & 4.12 & 4.04 & 4.6 & 4.16 & 3.5 \\
\hdashline
\multirow{4}{*}{GPT-3.5-Turbo} &
+ Instructor Knowledge & \textbf{4.56} & \textbf{4.82} & 4.68 & 4.36 & 3.76 \\
& + Student Knowledge & 3.92 & 4.18 & \textbf{4.8} & \textbf{4.46} & \underline{3.94} \\
& + \ours{} & \underline{4.3} & \underline{4.66} & \underline{4.7} & \underline{4.4} & \textbf{4.06} \\
\bottomrule
\end{tabular}
}
\caption{G-Eval result between YATA and other models. The best results for each base model are \textbf{bolded} and the second-best result is \underline{underlined}.}
\label{tab:exp}
\vspace{-10pt}
\end{table*}

%% file: 5_exp_new.tex
\section{Experiments}
To evaluate the efficacy of the \ours{} framework in achieving personalization for both instructors and students, we employ two complementary methods: 1) G-Eval~\citep{liu2023gpteval} to quantitatively assess YA-TA's responses across multiple criteria, and 2) case studies to qualitatively analyze the \ours{} framework.

\subsection{G-Eval}
\subsubsection{Experiment Setup}
\paragraph{Test set construction.} We generate the test set for evaluation by simulating a scenario where students with various academic backgrounds ask different questions about the lectures, and YA-TA provides answers to each question. We select an English course for computer science (CS50 from Harvard University\footnote{We obtained permission to use the lecture videos for this paper from the lecturer. The videos can be accessed at the following link: \url{https://www.youtube.com/watch?v=8mAITcNt710}}) as the testbed. Then, we extract potential questions from the lectures using \texttt{GPT-3.5-Turbo}. One of the authors, who is a CS expert, filters 10 high-quality questions from the questions extracted by \texttt{GPT-3.5-Turbo}. Additionally, we create profiles for 5 students with diverse majors and academic backgrounds. As each question is matched with multiple student profiles, we generate 50 test sets, each comprising a query, student knowledge, and instructor knowledge.

\paragraph{Baselines.} 
We set two baselines using GPT-3.5-Turbo and GPT-4o, where both are provided only the dialogue context without any retrieved knowledge.\footnote{In this section, we utilize \texttt{gpt-3.5-turbo-0125} and \texttt{gpt-4o-2024-05-13}.} Additionally, we conduct an ablation study to investigate the effect of each type of knowledge. These models are then instructed to generate responses to queries from the test set.

\subsubsection{Evaluation Criteria}
We employ G-Eval~\citep{liu2023gpteval} to assess performance across various criteria, scoring from 0 to 5. Instructor-side metrics are: (1) \textbf{Precision}: Does the answer provide necessary information without redundancy?; (2) \textbf{Groundedness}: Is the answer aligned with the instructor’s statements and teaching philosophy? Student-side metrics are: (1) \textbf{Helpfulness}: How satisfied is the student likely to be?; (2) \textbf{Comprehensiveness}: Does the answer appropriately consider the student's academic ability? Lastly, for both sides: (1) \textbf{Overall}: Does the response align with the instructor's statements and reflect the student's information?

\subsubsection{G-Eval Results}

Table~\ref{tab:exp} shows that retrieving information from just one side outperforms dual retrieval, which highlights the challenge of achieving personalization on both sides. Additionally, the highest performance achieved by the \ours{} framework when both sides are considered together demonstrates that our framework excels at integrating knowledge from both perspectives.

\input{tables/case_study_social_science}
\input{tables/case_study_computer_science}

\subsection{Case Study}
In this section, we present a qualitative analysis of our \ours{} framework through case studies of YA-TA interacting with a designed student.
\subsubsection{Experiment Setup}
\paragraph{Course setting.} We select the CS50 lecture used in G-Eval, along with a Korean course for social science (GPP6003 from Yonsei University)\footnote{For convenience, we named the course identifier GPP6003. The videos can be accessed at the following link: \url{https://www.learnus.org/local/ubonline/view.php?id=216527}}, as our test bed.

\paragraph{Student setting.} We set the virtual student as a third-semester graduate student majoring in Artificial Intelligence. Additionally, we assume this student has taken computer-related courses and achieved very high grades but has not taken any social science courses. Based on the student knowledge, we assume a high understanding level of computer science and a lower proficiency in social sciences.

\input{figures/figure_3}
\paragraph{Example of personalized response.}
As illustrated in Table~\ref{tab:case1} and Table~\ref{tab:case2}, YA-TA produces responses that are personalized for both the instructor and the student. The similarity between the evidence and YA-TA's responses shows that YA-TA bases its answers on the lecture. Furthermore, YA-TA uses examples relevant to the student's background, which demonstrates its ability to tailor responses based on the student's academic background.

\paragraph{Failure cases and limitation.}
Insufficient responses may result from the limitations of YA-TA, as it is designed to respond based on the instructor knowledge. When a student's question falls outside or only slightly overlaps with the lecture's scope, there may not be enough evidence to generate a helpful answer, leading to insufficient responses.

%% file: tables/case_study_social_science.tex
\begin{table}[ht!]
    \small 
    \begin{tabular}{p{\linewidth}}
    \toprule
        \textsc{\textbf{Social Science} (Yonsei GPP6003) } \\
    \midrule
        \info \space \textbf{Student Information} \\
        Major: Artificial Intelligence \\ Degree: Graduate Student \\ Semester: 3 \\ 
        Grade: (Data Structures, A), (Algorithms, A+), ... \\ 
    \midrule
        \query \space \textbf{Student's Query} \\
        Explain the relationship between capitalism and democracy. \\
    \midrule
        \evidence \space \textbf{Evidence} \\
        However, \textbf{it is generally considered difficult to maintain democracy in the absence of capitalism.} timestamp: (start: 819.96, end: 904.66) ... \\ 
    \midrule
        \YATAresponse \space \textbf{YA-TA Response} \\        Capitalism and democracy interact with each other. \\
        ... \\
       \textbf{capitalism plays an important role in the continuous maintenance and development of democracy.} \\
        ... \\
        In the field of artificial intelligence, this relationship is particularly evident. For example, in a capitalist market, \textbf{the advancement of AI technology is driven by competition, leading to innovation and economic growth}. \\
    \midrule
        \GPTresponse \space \textbf{GPT-4o Response} \\
        Capitalism and democracy are two important concepts in modern society, representing economic and political systems, respectively. \\
        ... \\
        The way capitalism and democracy are combined can vary widely depending on each country's historical and cultural context. \\
    \bottomrule
    \end{tabular}
    \caption{A sample translated response from YA-TA and GPT-4o.}
    \label{tab:case1}
\end{table}

%% file: tables/case_study_computer_science.tex
\begin{table}[ht!]
    \small
    \begin{tabular}{p{\linewidth}}
    \toprule
        \textsc{\textbf{Computer Science} (Harvard CS50)} \\
    \midrule
    \info \space \textbf{Student Information} \\
        Major: Artificial Intelligence \\ Degree: Graduate Student \\ Semester: 3 \\ 
        Grade: (Data Structures, A), (Algorithms, A+), ... \\ 
    \midrule
     \query \space \textbf{Student Query} \\
        Which sorting algorithm is described as having a time complexity of n log n. \\
    \midrule
        \evidence \space \textbf{Evidence} \\
        And then this one got a little mathy: log base 2 of n. And \textbf{log base 2 of n just means what is the value when you take n and divide it by two again and again.} timestamp:(start:1778.84, end:1790.4)  ... \\
    \midrule
       \YATAresponse \space \textbf{YA-TA Response} \\
        The sorting algorithm described as having a time complexity of n log n is merge sort. \\
        ... \\
        The process of dividing the array into halves and then merging them requires \textbf{log base 2 of n steps, as you divide the array size by two repeatedly.} \\
        ... \\
        \textbf{Given your background in Data Structures, Algorithms, and Machine Learning}, you would appreciate that merge sort is particularly useful for sorting large datasets efficiently. \\
    \midrule
        \GPTresponse \space \textbf{GPT-4o Response} \\
        Several sorting algorithms have a time complexity of $O(n log n)$. The most notable ones include:\\
        1. Merge Sort: ... \\
        2. Heapsort: ... \\
        3. Quicksort: ... \\
         These algorithms are widely used due to their efficiency and effectiveness in sorting large datasets. \\
    \bottomrule
    \end{tabular}
    \caption{A sample response from YA-TA and GPT-4o.}
\label{tab:case2}
\end{table}

%% file: figures/figure_3.tex
\begin{figure*}[ht!]
    \centering
    \includegraphics[width=0.98\linewidth]{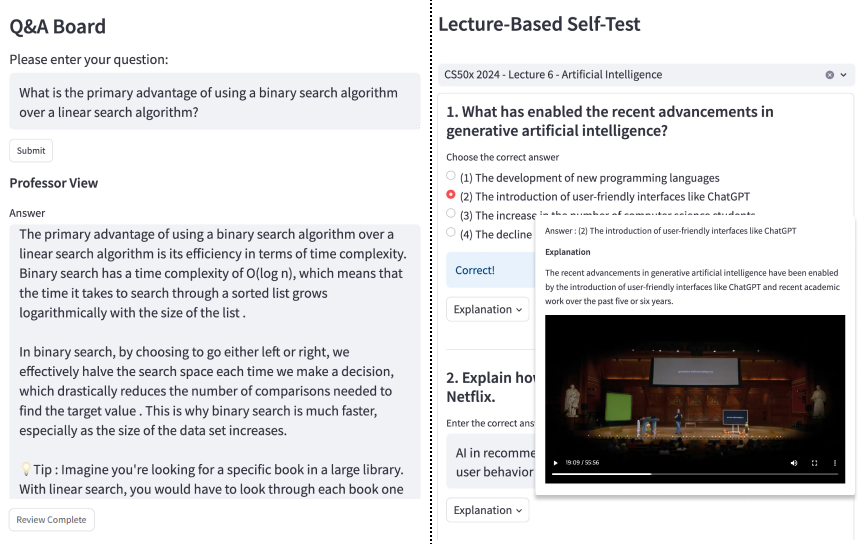}
    \caption{Extension of YA-TA}
    \label{fig:extension}
\end{figure*}

%% file: 5_results.tex
\section{Extension of YA-TA}
To enhance learning effectiveness, we provide two additional educational tools: Q\&A Board and Self-Practice. The Q\&A Board strengthens interaction between instructors and students, supporting in-depth learning. The Self-Practice tool enables students to test and review what they have learned.

\begin{itemize}
    \item \textbf{Q\&A Board:}
    The Q\&A board allows students to ask questions beyond the lecture and seek additional help. When a student posts a question, YA-TA drafts a response based on the instructor's knowledge. The instructor then reviews and refines this draft, ensuring it aligns with their teaching philosophy with minimal effort. This process facilitates direct student-instructor engagement while enabling instructors to efficiently provide thoughtful and personalized responses.

    \item \textbf{Self-Practice:} Our system also provides quizzes to allow students to self-assess their understanding of the instructor's knowledge. Quizzes are generated based on the instructor knowledge by focusing on the highlighted key points. Using a similar prompt as in the instructor-side retrieval process of YA-TA, quizzes help students evaluate their grasp of the material that the instructor considers important, thereby enhancing their learning experience.
\end{itemize}

%% file: 2_rel.tex
\section{Related Work}
Personalized LLMs have been extensively studied in NLP for educational purposes by implementing methods such as training datasets with instructor knowledge and assessing student academic levels~\citep{porsdam2023autogen, wozniak2024personalized}. Previous studies on instructor-side personalization involve fine-tuning models with specific datasets to generate customized responses~\citep{hicke2023chata, chevalier2024language, li2023curriculum, macina2023mathdial, chae2023tutoring}, and using Retrieval-augmented Generation~\citep{levonian2023retrieval}. Student-side personalization tailors learning experiences to individual academic levels, with systems that dynamically adjust to student needs~\citep{chen2023empowering} and offer personalized learning paths~\citep{sajja2023artificial}.

%% file: 6_conc.tex
\section{Conclusion}

In this study, we propose YA-TA, a multi-turn QA agent that provides personalized responses for both instructors and students. YA-TA can be set up with just a lecture video and a single click, without the need for additional model training, making it highly versatile. Additionally, our platform enhances the educational experience by offering extensions such as Q\&A Board and Self-Practice.

\section{Limitation}

While YA-TA offers extensive services to enhance the learning experience, several limitations must be acknowledged.

Firstly, when extracting knowledge from both instructors and students, there is a potential risk of patent and privacy issues. It is crucial to obtain explicit permission from instructors before using their videos and to secure consent from students when using their personal information. Additionally, the reliance on OpenAI's API may lead to cost-related challenges.

Moreover, since we build memory specific to each lecture and student, our framework cannot integrate content from multiple courses into a single response. Our approach also does not account for the diverse features a student may exhibit across different courses, as it analyzes queries posed by the student within a single course rather than across multiple courses.

\section{Acknowledgments}
This work was supported by Institute of Information \& Communications Technology Planning \& Evaluation (IITP) grant funded by the Korean government (MSIT)(No.RS-2020-II201361, Artificial Intelligence Graduate School Program (Yonsei University)) and (No.RS-2021-II212068, Artificial Intelligence Innovation Hub) and (No.RS-2022-II220077,AI Technology Development for Commonsense Extraction, Reasoning, and Inference from Heterogeneous Data). Jinyoung Yeo and Dongha Lee are the co-corresponding authors
(jinyeo@yonsei.ac.kr, donalee@yonsei.ac.kr).

%% file: 8_appendix.tex
\section{YA-TA Details}
\label{sec:appendix_yata_details}
We use BM25 ~\citep{robertson2009probabilistic} for the sparse retriever and \texttt{text-embedding-ada-002} for the dense retriever in the instructor knowledge retrieval phase. Also, we set the number of retrieved segments $k$ as 10. For both LLM reasonser and response generator, we use \texttt{gpt-3.5-turbo-0125}.

\section{G-Eval Details}
\label{sec:appendix_geval}
\subsection{Student Information}
We aim to evaluate YA-TA's ability to achieve student-side personalization for various students. To meet this end, we create virtual student profiles with diverse academic backgrounds.
\input{figures/geval/student_info}
\subsection{Questions}
First, we generate advocate questions likely to arise from the CS50 course using the prompt employed to create the quiz. Then, a CS expert selects 10 high-quality questions that students would ask.
\input{tables/questions}
\subsection{G-Eval Criteria}
We utilize G-Eval prompts to assess 5 criteria. Figure~\ref{fig:precision} shows the example prompt we used for G-Eval.
\input{figures/geval/precision}

%% file: figures/geval/student_info.tex
\begin{figure}[ht!]
    \centering
    \includegraphics[width=0.98\linewidth]{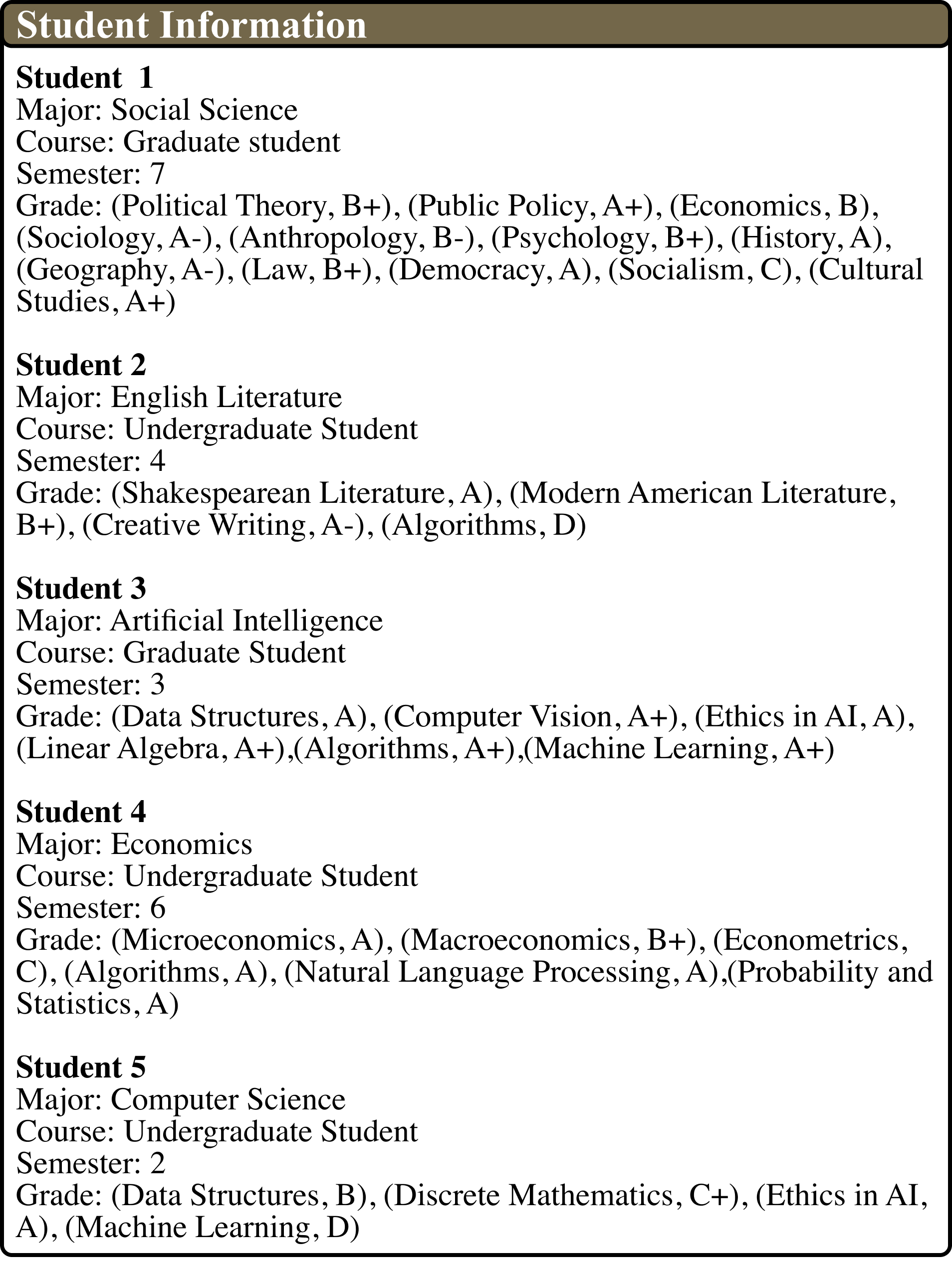}
    \caption{Virtual student profiles that we set}
    \label{fig:student_info}
\end{figure}

%% file: tables/questions.tex
\begin{table}[ht!]
\centering
\begin{tabular}{p{\dimexpr\linewidth-2\tabcolsep}} 
\toprule
\textbf{Question} \\
\midrule
1. How does AI improve the experience of using recommendation systems in services like Netflix? \\
2. What is the primary purpose of "rubber duck debugging" in programming? \\
3. Which sorting algorithm is described as having a time complexity of n log n in the lecture? \\
4. What is the purpose of the base case in a recursive function? \\
5. What is the difference between a user prompt and a system prompt in the context of AI-based educational tools? \\
6. What is the primary difference between traditional AI approaches and machine learning in the context of solving games? \\
7. Describe the basic algorithm for playing the game Breakout as explained in the lecture. \\
8. What real-world example is used to explain the divide and conquer algorithm in the lecture? \\
9. What is the primary advantage of using a binary search algorithm over a linear search algorithm? \\
10. What is the main reason for using pseudocode before writing actual code? \\
\bottomrule
\end{tabular}
\caption{10 high-quality questions from the questions extracted by GPT-3.5-Turbo.}
\label{tab:questions}
\vspace{-10pt}
\end{table}

%% file: figures/geval/precision.tex
\begin{figure}[ht!]
    \centering
    \includegraphics[width=0.98\linewidth]{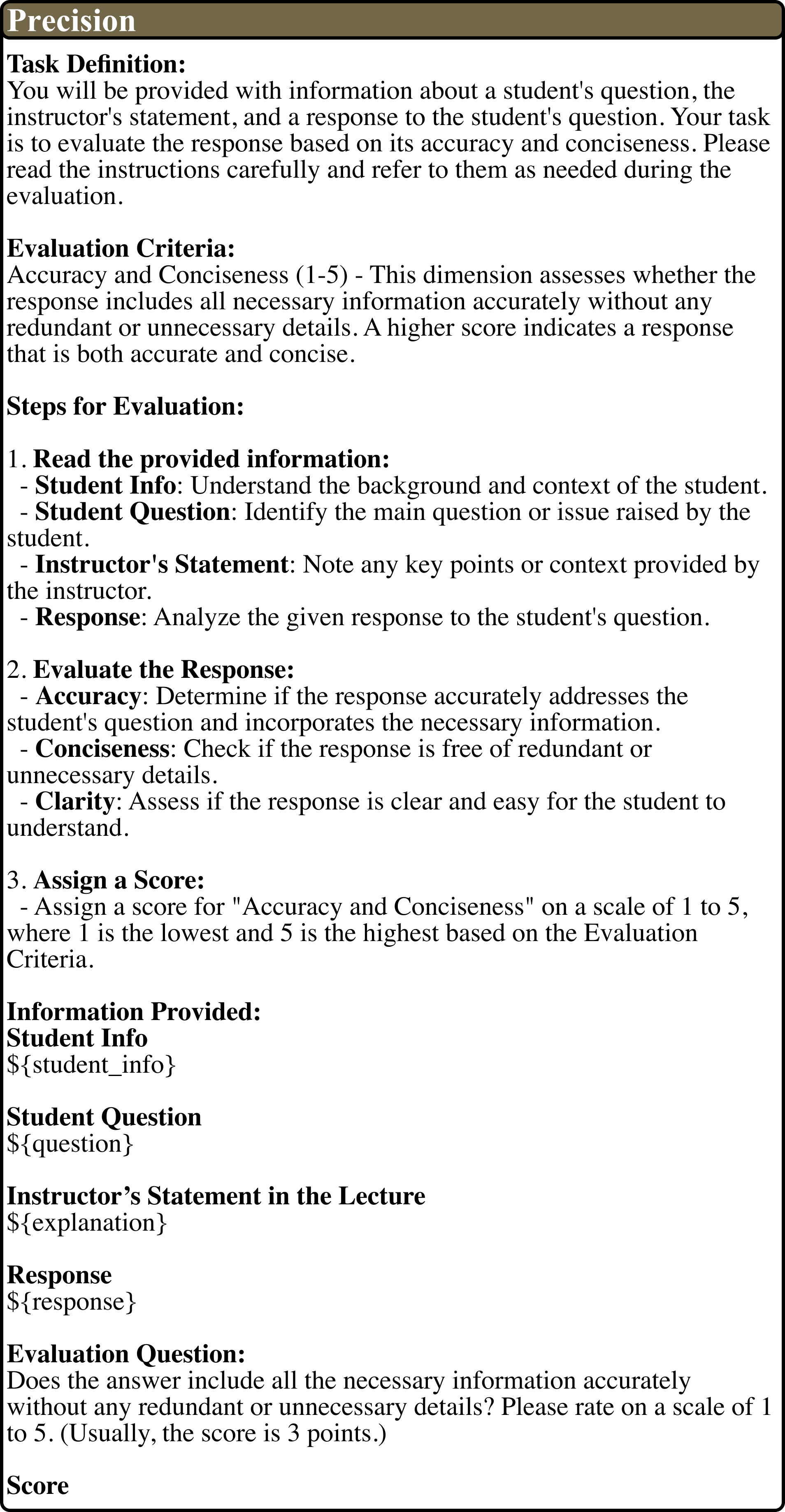}
    \caption{G-Eval prompt used to assess precision}
    \label{fig:precision}
\end{figure}